\documentclass[10pt,twocolumn,letterpaper]{article}

\usepackage{cvpr}              

\usepackage{multirow}
\usepackage{ifthen}
\usepackage{graphicx}
\usepackage{algorithm}
\usepackage{algorithmic}
\usepackage{comment}
\usepackage[table]{xcolor}

\usepackage{xcolor}
\usepackage{amsmath}
\usepackage{amssymb}
\usepackage[table]{xcolor}
\usepackage{threeparttable} 
\usepackage[normalem]{ulem}
\usepackage{wrapfig}
\usepackage{adjustbox}
\usepackage[table]{xcolor}

\usepackage{xcolor}

\newboolean{K_is_15}

\setboolean{K_is_15}{false}

\definecolor{fftcolor}{RGB}{210, 10, 0}     
\definecolor{loracolor}{RGB}{0, 80, 180}    

\definecolor{cvprblue}{rgb}{0.21,0.49,0.74}
\usepackage[pagebackref,breaklinks,colorlinks,allcolors=cvprblue]{hyperref}

\def\paperID{XXX} 
\def\confName{CVPR}
\def\confYear{2027}

\title{Exploring Motion-Language Alignment for Text-Driven Motion Generation}

\author{
Ruxi Gu$^{1,2}$ \quad Zilei Wang$^{1*}$ \quad Wei Wang$^{2,3}$\thanks{Corresponding authors. Emails: \texttt{zlwang@ustc.edu.cn}, \texttt{wangwei@ustb.edu.cn}.}\\
$^{1}$Department of Automation, University of Science and Technology of China\\
$^{2}$State Key Laboratory of General Artificial Intelligence, BIGAI\\
$^{3}$School of Computer and Communication Engineering, University of Science and Technology Beijing
}

\begin{document}
\maketitle
\begin{abstract}
Text-driven human motion generation aims to synthesize realistic motion sequences that follow textual descriptions. Despite recent advances, accurately aligning motion dynamics with textual semantics remains a fundamental challenge. In this paper, we revisit text-to-motion generation from the perspective of motion-language alignment and propose MLA-Gen, a framework that integrates global motion priors with fine-grained local conditioning. This design enables the model to capture common motion patterns, while establishing detailed alignment between texts and motions. Furthermore, we identify a previously overlooked attention sink phenomenon in human motion generation, where attention disproportionately concentrates on the start text token, limiting the utilization of informative textual cues and leading to degraded semantic grounding. To analyze this issue, we introduce SinkRatio, a metric for measuring attention concentration, and develop alignment-aware masking and control strategies to regulate attention during generation. Extensive experiments demonstrate that our approach consistently improves both motion quality and motion-language alignment over strong baselines.
\end{abstract} 

\section{Introduction}

Human motion generation from natural language \citep{tevet2022human, meng2025rethinking} aims to synthesize realistic motion sequences that faithfully follow textual descriptions. This task has attracted increasing attention due to its wide-ranging applications in character animation, virtual reality, and human-robot interaction. Recent approaches have made substantial progress by leveraging large-scale motion datasets \citep{guo2022generating, mahmood2019amass, guo2020action2motion} together with powerful generative models, including diffusion-based \citep{zhang2024motiondiffuse, zhang2023remodiffuse} and flow-based \citep{lipman2024flow, meng2025absolute} models.

\begin{figure}[t]
\centering
\includegraphics[width=0.45\textwidth]{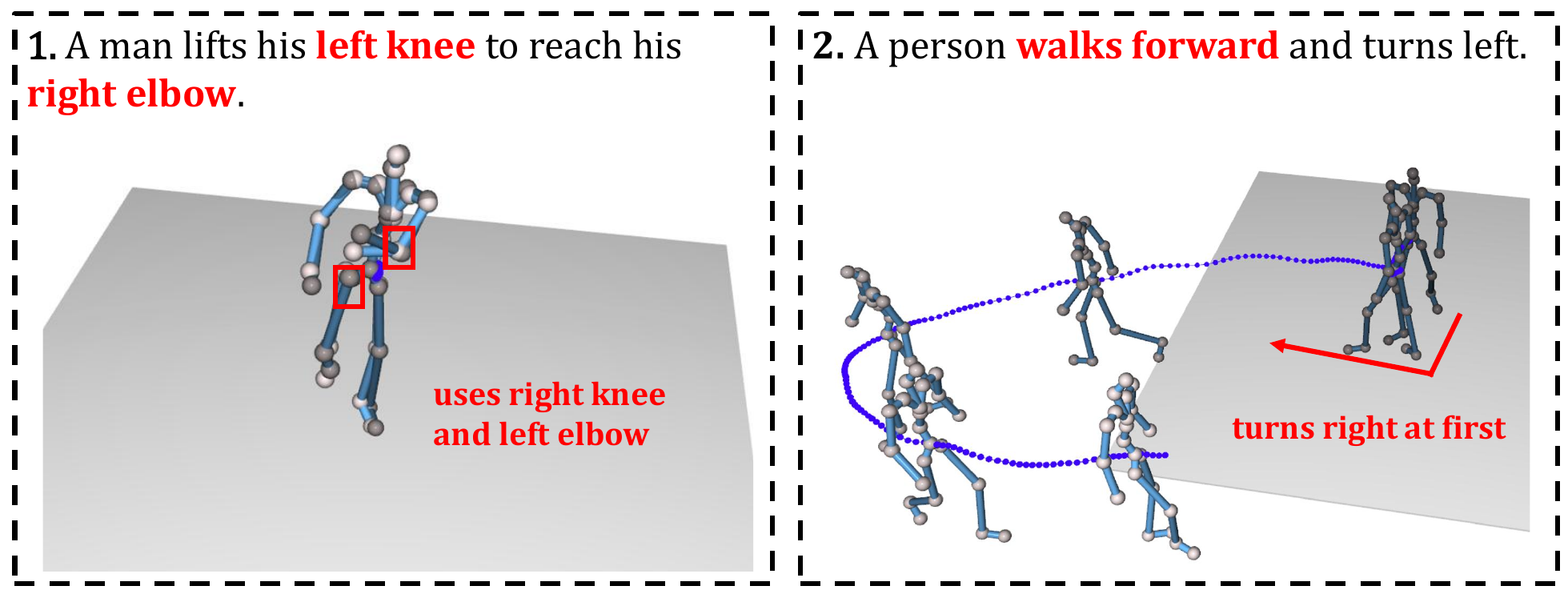}
\caption{Failure cases from previous text-to-motion generation framework \cite{meng2025absolute}. The color gradient from dark to light represents the temporal flow.
}
\label{fig:old_framework}
\end{figure}

Despite these advances, achieving precise motion-language alignment remains a fundamental challenge. Existing methods \citep{tevet2022human, guo2022generating, meng2025absolute} typically rely on global text representations derived from pretrained CLIP models \citep{radford2021learning} to guide motion generation. While effective at capturing coarse semantics, such global conditioning provides limited correspondence between individual textual concepts and motion dynamics over time. Consequently, generated motions may capture the overall intent of a description while failing to accurately reflect fine-grained semantic cues, as illustrated in Fig.~\ref{fig:old_framework}. This observation motivates us to revisit text-to-motion generation from the perspective of \textbf{motion-language alignment}: how can we establish more precise correspondence between textual concepts and motion dynamics during generation?

To address this question, we propose \textbf{MLA-Gen}, a text-driven motion generation framework that explicitly explores motion-language alignment from complementary perspectives. First, effective fine-grained alignment requires a structured motion-side representation that captures recurring motion patterns and provides meaningful features for semantic grounding. We therefore introduce a set of learnable \emph{memory slots} that capture global motion prototypes shared across sequences, providing a compact representation of recurring motion patterns and enhancing structural consistency. Building on this motion representation, we introduce a \emph{local fine-grained alignment} mechanism that performs cross-attention between motion frames and text tokens, explicitly establishing correspondence between textual concepts and motion dynamics over time.

\begin{figure*}[t]
\centering
\includegraphics[width=0.75\textwidth]{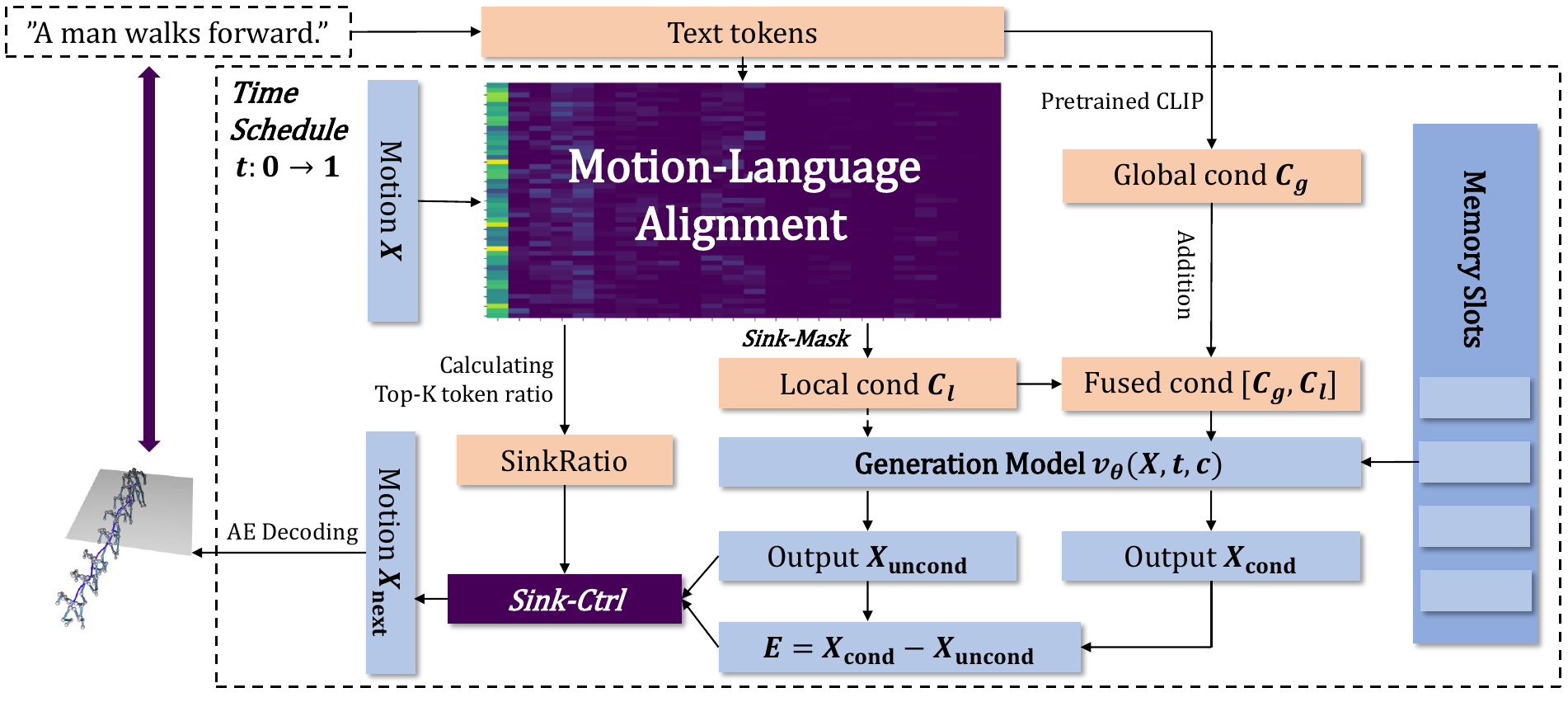}
\caption{Overview of our MLA-Gen framework. It comprises three complementary components: Memory Slots for capturing global motion priors, Motion-Language Alignment for providing fine-grained textual semantics, and a SinkRatio-based mechanism that models and mitigates the attention sink during both attention computation (\textit{sink-mask}) and sampling (\textit{sink-ctrl}).}
\label{fig:new_framework}
\end{figure*}

Beyond explicitly modeling fine-grained correspondence, we further investigate how the alignment process behaves during generation. We observe a phenomenon analogous to the \emph{attention sink} effect reported in transformer-based language models \citep{barbero2025llms, xiao2023efficient}, where cross-attention weights disproportionately concentrate on the start token of the text sequence. Such concentration can divert attention from semantically informative tokens and weaken fine-grained motion-language correspondence. To quantify this behavior, we introduce \textit{SinkRatio}, a metric that measures the degree of attention concentration on the start token. Based on this observation, we develop two complementary strategies to mitigate the sink effect. First, \emph{sink-mask} suppresses the dominance of the start token at selected timesteps, encouraging the model to attend to more informative textual tokens. Second, inspired by CFG-ctrl \citep{wang2026cfg}, \emph{sink-ctrl} introduces an alignment-aware Classifier-Free Guidance (CFG) mechanism that adaptively adjusts the guidance strength according to SinkRatio, improving semantic alignment during sampling. An overview of MLA-Gen is illustrated in Fig.~\ref{fig:new_framework}.

Extensive experiments demonstrate the effectiveness of MLA-Gen in both motion generation and motion-language alignment. Compared with the existing text-to-motion generation framework, MLA-Gen substantially improves generation quality, reducing FID from $0.107$ to $0.056$ for the small-scale model and from $0.083$ to $0.040$ for the big-scale model. Qualitative visualizations further show that MLA-Gen captures finer human-body details while preserving stronger temporal consistency in generated motion sequences.
 
Our contributions can be summarized as follows:
\begin{itemize}
    \item (1) We propose \textbf{MLA-Gen}, a text-driven motion generation framework that explores motion-language alignment through motion-side memory representations and fine-grained cross-modal attention.
    \item (2) We identify an attention sink phenomenon in motion-language alignment and introduce \textit{SinkRatio} to quantify attention concentration. Based on this observation, we develop \emph{sink-mask} and \emph{sink-ctrl} to improve fine-grained semantic alignment during generation.
    \item (3) Extensive experiments demonstrate that MLA-Gen improves both motion generation quality and motion-language alignment, with substantial gains in quantitative and qualitative evaluations.
\end{itemize}

\section{Related Works}

\subsection{Human Motion Generation}

Existing approaches to human motion generation generally differ in how human motion is represented and generated, with representations broadly categorized as either continuous or discrete. For continuous motion representations, recent methods predominantly adopt diffusion-based or flow-based generative models. Some approaches operate directly in the pose-frame space \citep{tevet2022human, shafir2023human, dabral2023mofusion, zhang2023remodiffuse, zhang2024large, zhou2024emdm, chen2024taming,  zhang2024motiondiffuse, li2025unimotion,   chen2025free}. Although these models have demonstrated strong capabilities in producing realistic and diverse motions, directly modeling the pose-frame space can be sensitive to noise in motion capture datasets, which may introduce artifacts in synthesized motions. To address this limitation, latent diffusion approaches \citep{ dai2024motionlcm, meng2025rethinking, meng2025absolute, tuautoregressive, xiao2025motionstreamer, zhang2025flashmo, zhang2024motion, jiang2025motionpcm} first encode motion sequences into a compact latent space before generation. This strategy can effectively improve stability and efficiency of the generation process; however, compressing the entire sequence that contains detailed temporal dynamics into a motion latent representation may lead to the loss of fine-grained motion details.

Another line of research \citep{guo2022tm2t, zhang2023generating, jiang2023motiongpt,guo2024momask,  ghosh2025duetgen, hwang2025snapmogen, hong2025egolm,  liu2025gesturelsm, wang2025motiondreamer,  zhang2025kinmo} adopts discrete motion representations through vector quantization \citep{van2017neural}. In these methods, continuous motion sequences are first mapped to discrete codebook tokens and then modeled using a next-token prediction paradigm. 
The discretization process often preserves temporal structure and enables more efficient training. Nevertheless, converting continuous motion signals into a finite codebook inevitably introduces quantization losses. Recently, inspired by progresses in large-scale image generation models \citep{li2024autoregressive}, several works \citep{meng2025rethinking, tuautoregressive, xiao2025motionstreamer, zhu2025motiongpt3, he2025molingo} have explored continuous latent representations within auto-regressive generation frameworks. By predicting continuous-valued latent variables instead of discrete tokens, these approaches aim to combine the benefits of auto-regressive modeling and continuous representations. However, this also exacerbates the accumulation of errors and poses challenges to training stability.

MLA-Gen is implemented on a flow-based model; however, due to its modular and transferable design, along with the ubiquity of the attention sink phenomenon in both flow-based and auto-regressive models, it can be easily adapted to auto-regressive models.

\subsection{Attention Sink}

Attention sink refers to tokens that receive disproportionately large attention weights despite carrying little semantic information \citep{xiao2023efficient, barbero2025llms}. In auto-regressive large language models (LLMs) \citep{barbero2025llms, gu2024attention}, due to their sequential generation process, the model’s attention tends to concentrate on the start token of the text. In contrast, in diffusion-based language models (DLMs) \citep{rulli2025attention, wang2025sparsed, song2025sparse}, the diffusion steps are bidirectional and iterative, so the corresponding sink tokens can appear not only at the beginning but also at other masked or semantically neutral positions within the text. Recent studies \citep{barbero2025llms, gu2024attention} leverage the attention sink mechanism to further concentrate attention weights, mitigate excessive information mixing in long-context scenarios, while others \citep{wang2025sparsed, song2025sparse} design sparse attention patterns based on sink tokens to accelerate inference.

In LLMs, mitigating the attention sink, such as dropping the sink token’s weights during inference, can impair model performance \citep{barbero2025llms}. Similarly, we observe a comparable attention sink phenomenon in our flow-based motion generation model. Unlike prior findings in LLMs, reducing the attention sink here can partially encourage the model to attend to other semantically relevant text tokens.

\section{Method}
\subsection{Problem Formulation}

\noindent{\textbf{Motion representation.}}
We represent a motion sequence with $F$ frames and $J$ joints in absolute world coordinates as $Mot \in \mathbb{R}^{F \times J \times 3}$, where each joint is represented by its 3D position. Following the practice in \citep{meng2025absolute}, we encode the raw motion sequence into a latent representation using a pretrained motion autoencoder (AE) \citep{kingma2013auto}. The encoder compresses the motion into $X \in \mathbb{R}^{L \times J \times D_{ae}}$, where $L$ denotes the temporally down-sampled sequence length and $D_{ae}$ denotes the latent feature dimension (for brevity, we treat the down-sampled $L$ as a generalized motion-frame dimension based on the semantic similarity between adjacent frames). The decoder reconstructs the motion sequence  from the generated latent representation. In this work, the AE is used as a fixed backbone and trained independently with a standard reconstruction objective. 

\noindent{\textbf{Flow-based text-to-motion generation.}}
Given a textual description $y$, the goal of text-to-motion generation is to model the conditional distribution $p(X \mid y)$ in the latent motion space. We adopt a flow-based generative framework \citep{albergo2022building, lipman2022flow, liu2022flow} to model this distribution. Specifically, we learn a time-dependent velocity field $v_{\theta}(X,t,y)$ that transports samples from a simple prior distribution $p_0(X)$ (e.g., standard Gaussian) to the motion data distribution $q(X)$ conditioned on the text $y$.

Let $\psi_t : \mathbb{R}^{L \times J \times D_{ae}} \rightarrow \mathbb{R}^{L \times J \times D_{ae}}$ denote the flow map defined by the ordinary differential equation (ODE):
\begin{equation}
\frac{d\psi_t(X_0)}{dt} = v_{\theta}(\psi_t(X_0), t, y), 
\quad
\psi_0(X_0) = X_0,
\end{equation}
where $X_0 \sim p_0$ is the initial noise sample. Following the Rectified Flow formulation \citep{lipman2024flow, liu2022flow}, we consider a linear interpolation path between a noise sample $X_0$ and a data sample $X_1$:
\begin{equation}
X_t = (1-t)X_0 + t X_1, \quad t \in [0,1].
\end{equation}
Along this path, the target velocity field is constant and equal to $X_1 - X_0$. Our model is trained using the conditional flow-matching objective:
\begin{equation}
\mathcal{L}(\theta) =
\mathbb{E}_{
\substack{
t \sim \mathcal{U}(0,1), \\
X_0 \sim p_0, \\
X_1 \sim q
}
}
\left[
\left\| v_{\theta}(X_t,t,y)-(X_1-X_0) \right\|_2^2
\right].
\end{equation}

At inference, a noise $X_0 \sim p_0$ is sampled and propagated through the learned ODE from $t=0$ to $t=1$, producing the final motion latent  $\hat{X}_1 = \psi_1(X_0)$. The final motion sequence $\hat{M}$ is obtained by decoding $\hat{X}_1$ with the AE decoder.

\subsection{Motion-Language Alignment Modeling}

A key challenge in text-driven motion generation is to effectively align textual semantics with motion dynamics over time while maintaining coherent motion structure. To this end, we explicitly model motion-language alignment from two perspectives: (1) capturing global motion priors shared across sequences, and (2) establishing fine-grained temporal alignment between motion frames and textual tokens. 

For a given motion latent sequence $X \in \mathbb{R}^{L \times J \times D_{ae}}$ and its corresponding text $y$ at a certain timestep, we denote $C_g \in \mathbb{R}^{D_{clip}}$ as the global CLIP feature of $y$, and $h  \in \mathbb{R}^{L\times D_{flow}}$ as the hidden representation of $X$ within the flow model. 

During alignment, we first extract the frame-level representation $z \in \mathbb{R}^{L \times D_{flow}}$ of the current $X$ by encoding then averaging over the joint dimension, then obtain the token-level representation $T \in \mathbb{R}^{N \times D_{clip}}$ of $y$ via CLIP, where $N$ denotes the number of text tokens. Since $D_{flow}$ may exceed $D_{clip}$, we introduce two linear layers $W_\text{up} \in \mathbb{R}^{D_{flow} \times D_{clip}}$ and $W_\text{down} \in \mathbb{R}^{D_{clip} \times D_{flow}}$ to perform dimension adjustment in the alignment computation.

\noindent{\textbf{Global motion prior via memory slots.}} Human motions exhibit strong structural regularities across different actions. To capture such shared motion patterns, we introduce a set of learnable memory slots that act as global motion prototypes. Specifically, we maintain a memory matrix $M \in \mathbb{R}^{S \times D_{flow}}$, where $S$ denotes the number of slots. These slots are randomly initialized and optimized jointly with the model parameters during training.

\begin{figure}[t]
\centering
\includegraphics[width=0.45\textwidth]{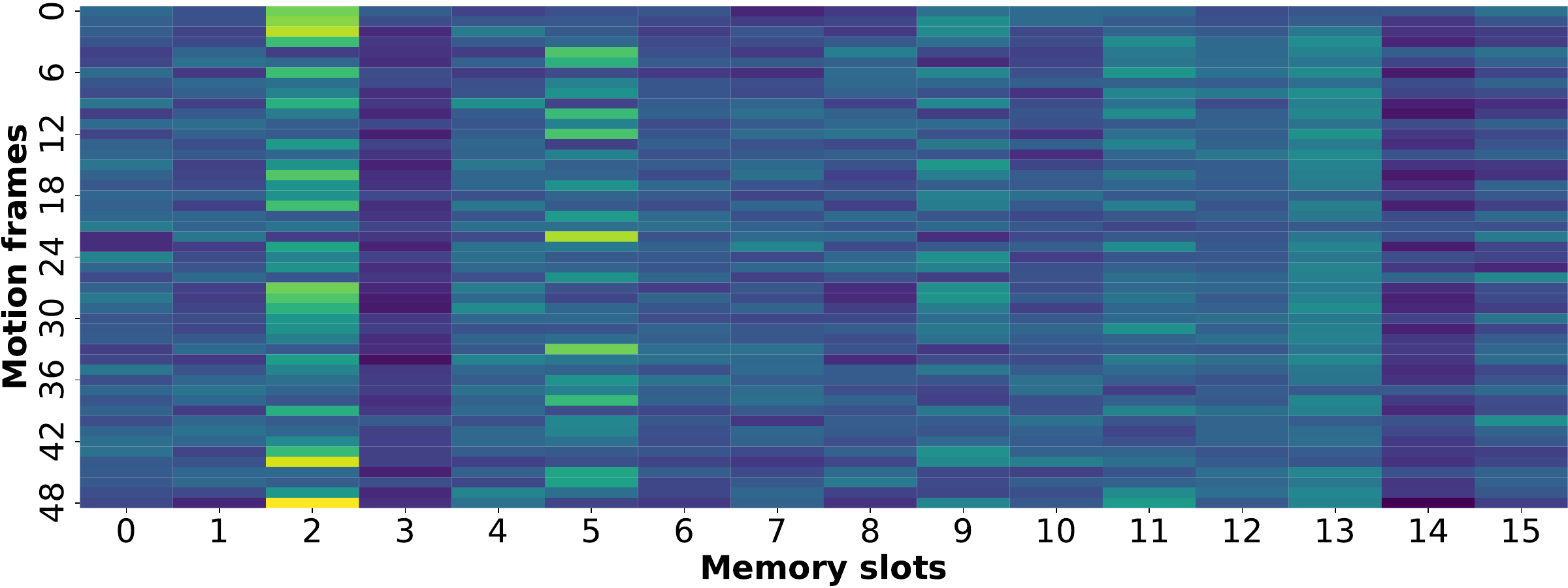}
\caption{Heatmap of the memory slots activation. Regions rendered in brighter yellow indicate higher attention weights between the corresponding motion frames and memory slots. 
}
\label{fig:heatmap_slots}
\end{figure}

Within each transformer layer of the flow model, we augment the hidden representation $h$ with a memory attention module. Here, the hidden features serve as queries, while the memory matrix provides keys and values:
\begin{equation}
\hat{h} = h + \text{Attn}(Q = h,\; K = M,\; V = M),
\end{equation}
where $\text{Attn}(\cdot)$ denotes the multi-head attention. Through this attention-based integration, hidden features can retrieve relevant global motion patterns from the memory slots. As illustrated in Fig. \ref{fig:heatmap_slots}, memory slots exhibit varying attention intensities, reflecting heterogeneous focus across slots and suggesting the presence of semantically distinct motion prototypes.

\noindent{\textbf{Local motion-language alignment. }}While global text embeddings $C_g$ provide coarse guidance, they often fail to capture the detailed textual semantics. To address this limitation, we introduce a local conditioning mechanism that enables fine-grained alignment between motions and texts.

Given the frame-level motion representation $z$ and the token-level text embedding $T$, we compute a cross-modal attention to obtain fine-grained textual conditions:
\begin{equation}
C_l = \text{Attn}(Q = z,\; K = W_\text{up}T,\; V = W_\text{up}T),
\end{equation}
where $C_l \in \mathbb{R}^{L \times D_{flow}}$. This allows each motion frame to dynamically aggregate relevant text semantics, establishing granular alignment along the temporal dimension.

\begin{figure}[h]
\centering
\includegraphics[width=0.45\textwidth]{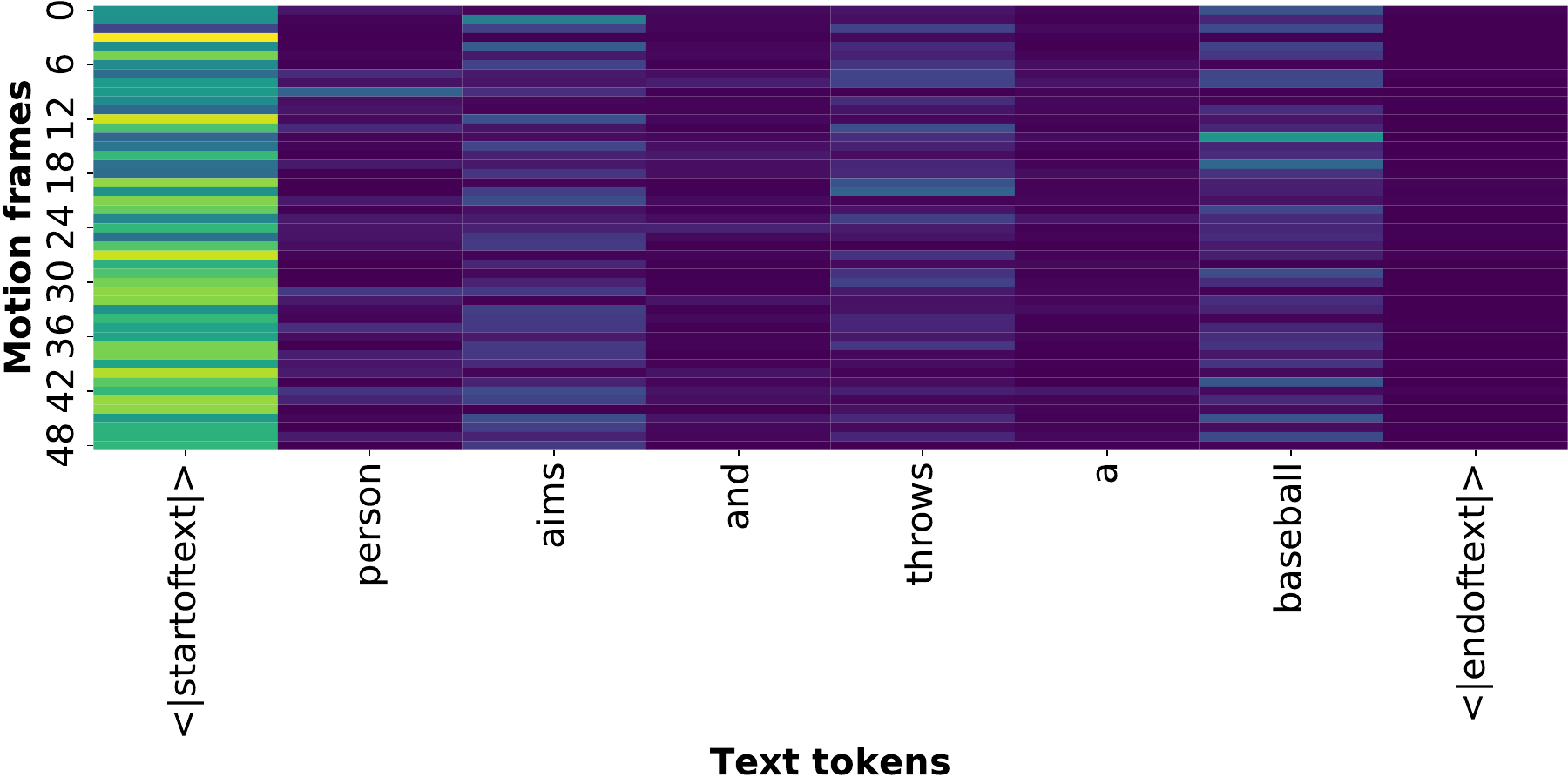}
\caption{Heatmap of motion-language alignment. Regions in brighter yellow indicate higher attention weights between the corresponding motion frames and text tokens.
}
\label{fig:heatmap}
\end{figure}

To incorporate both global and local textual information, we further fuse $C_g$ with the local condition via a weighted addition: $C = C_g + \lambda \cdot W_\text{down} C_l$, where $C_g $ is repeated along the $L$ dimension, and $\lambda$ is a fixed scalar controlling the contribution of local alignment. The aggregated condition $C \in \mathbb{R}^{L\times D_{clip}}$ is used to modulate the velocity network $v_{\theta}$ in the flow model. Fig. \ref{fig:heatmap} presents an example heatmap of the motion-language alignment. As can be observed, MLA-Gen focuses on text tokens that convey richer semantics, such as \texttt{<aims>}, \texttt{<throws>}, and \texttt{<baseball>}.

Similarly, MoLingo \citep{he2025molingo} also emphasizes alignment between text and motion. However, unlike MoLingo, which relies on additional frame-level supervised motion data for training, MLA-Gen leverages memory slots and local motion-language alignment without any corresponding supervision signals. This demonstrates that MLA-Gen can spontaneously achieve alignment without external knowledge, and can be readily extended to other domains, such as human-human interaction generation and human-object interaction generation.

\begin{figure*}[h]
\centering
\includegraphics[width=0.45\textwidth]{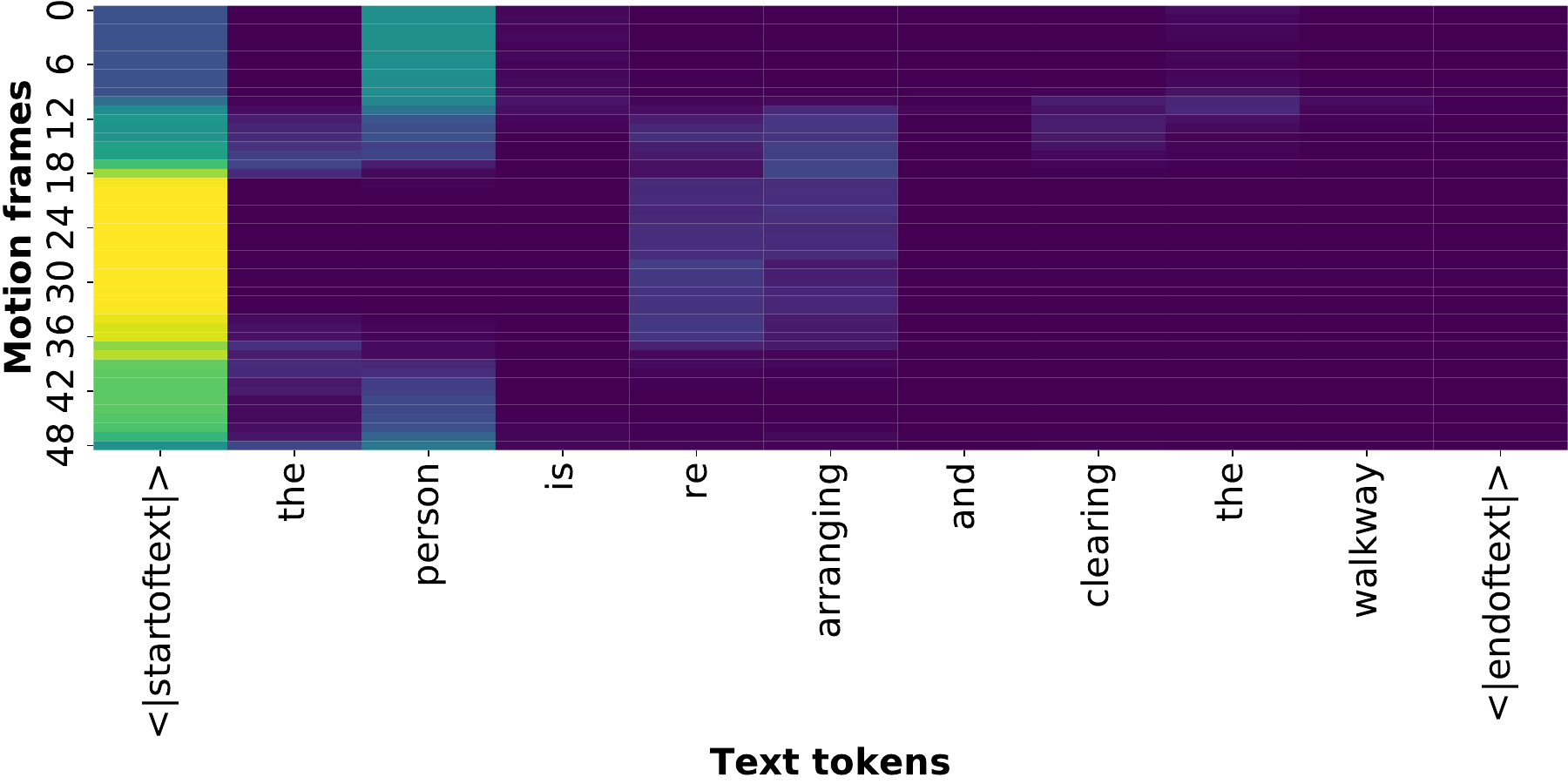}
\hspace{0.01\textwidth}
\includegraphics[width=0.45\textwidth]{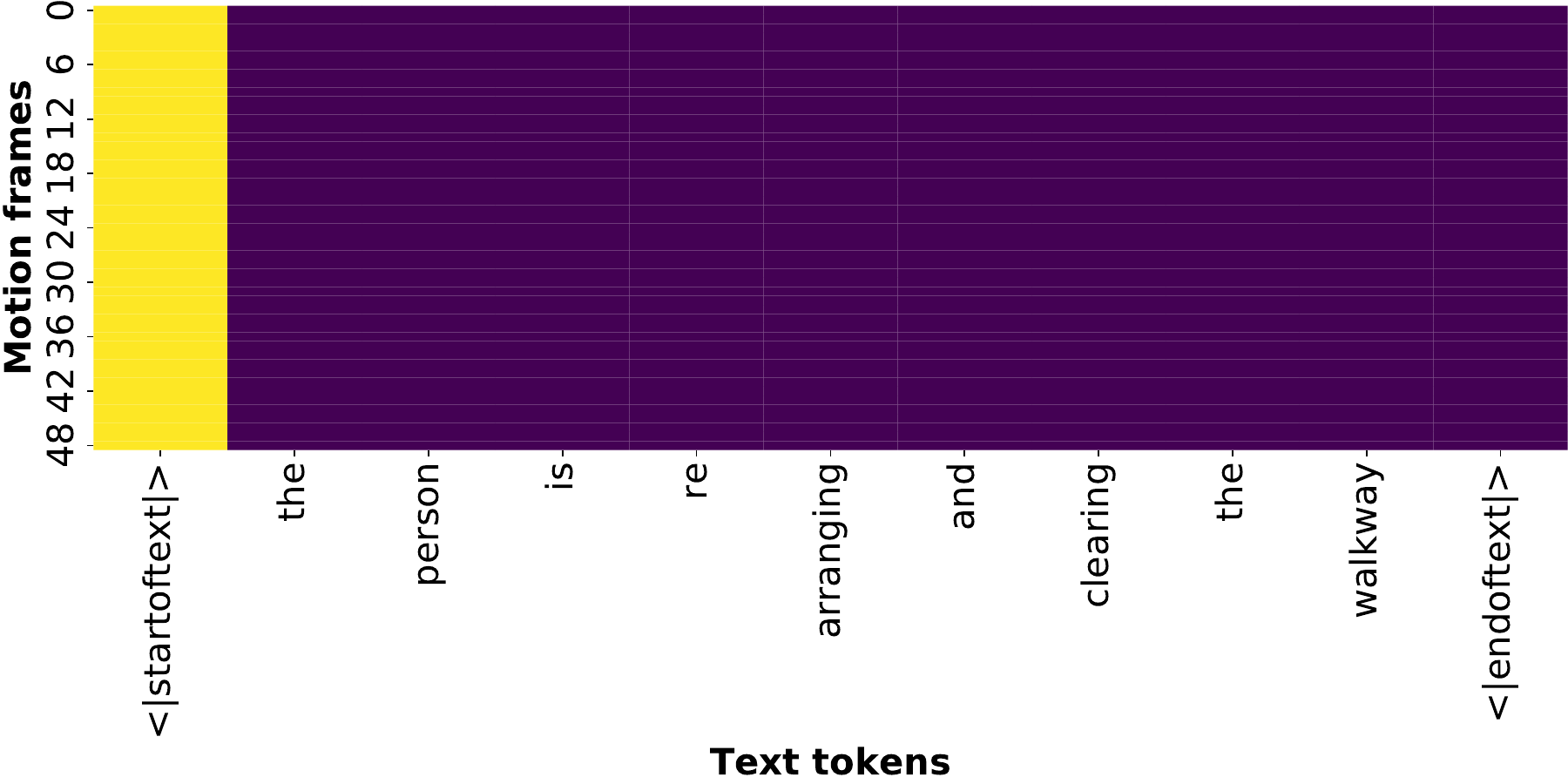} 

\caption{Heatmaps comparison of alignment on the masked model (left) and the unmasked model (right). 
 }
\label{fig:heatmap_compare}
\end{figure*}

\subsection{Attention Sink in Motion-Language Alignment}
\label{Sec3.3}

In local motion-language alignment, we observe a systematic bias in the resulting cross-modal attention patterns: the attention weights consistently concentrate excessively on the first token of the text sequence, as demonstrated in Fig. \ref{fig:heatmap}. This phenomenon resembles the \emph{attention sink} behavior previously reported in transformer-based language models \citep{barbero2025llms, rulli2025attention}, where a small subset of tokens absorb a disproportionate amount of attention mass.

In the context of motion generation, this phenomenon can negatively affect the utilization of textual semantics. When attention overly focuses on \texttt{<start token>}, which carries no explicit meaning, the model may rely primarily on a coarse global semantic anchor while under-utilizing the remaining tokens that contain richer information. As a result, the generated motion may match the overall meaning but fail to capture subtle dynamic details.

\noindent{\textbf{SinkRatio: Quantifying attention sink.}} To better analyze this phenomenon, we introduce a metric called \emph{SinkRatio} to measure the degree of attention concentration in motion-language alignment.

Let $A \in \mathbb{R}^{L \times N}$ denote the cross-attention matrix between motion frames and text tokens, where $L$ is the number of motion frames and $N$ is the number of text tokens. Each element $A_{i,j}$ represents the attention weight assigned from the $i$-th frame to the $j$-th token. For each motion frame, we select the top-$K$ largest attention weights and compute their sum $s_i = \sum_{k \in \text{Top-}K(A_i)} A_{i,k}$. The SinkRatio is then defined as the average concentration across all frames:
\begin{equation}
\text{SinkRatio} = \frac{1}{L} \sum_{i=1}^{L} s_i.
\end{equation}
A higher SinkRatio indicates that the attention distribution is more concentrated on a small subset of text tokens, reflecting a stronger attention sink effect. Conversely, a lower SinkRatio suggests a more evenly distributed attention pattern and better utilization of textual information.

We adopt a top-$K$ strategy to quantify the attention sink, because during inference, the largest attention weights are consistently assigned to the leading text tokens, which typically contain little motion-specific semantics. This prevents erroneously attributing overly high importance to more informative tokens.

In practice, we find that after masking the attention weights of \texttt{<start token>}, although the attention distribution becomes more balanced, the sink phenomenon can still occur at the subsequent positions. According to the analysis in prior works \cite{barbero2025llms, xiao2023efficient}, this behavior may reflect an adaptive characteristic of the model itself. Subsequently, we leverage this observation to design \textbf{sink-aware generation strategies} that mitigate attention bias and improve motion-language alignment.

\subsection{Sink-aware Motion Generation}

Building on SinkRatio, we design two components that explicitly mitigate attention sink bias: \emph{sink-mask}, a sink-aware token weight masking strategy that encourages more balanced cross-modal attention, and \emph{sink-ctrl}, an adaptive classifier-free guidance mechanism that adaptively regulates conditional guidance.

\noindent{\textbf{\textit{Sink-mask}: Sink-aware token masking.}} As discussed in Section \ref{Sec3.3}, cross-modal attention between motions and texts often exhibits a strong bias toward the \texttt{<start token>}. While serving as a stable global semantic anchor, over-reliance on this token limits the model’s ability to utilize detailed information from other textual tokens.

To alleviate this problem, we introduce \emph{sink-mask}, a sink-aware token masking strategy applied during both training and generation. The key idea is to diminish the influence of \texttt{<start token>} in the cross-attention computation at later sampling rounds, thereby encouraging the model to attend to a broader set of text tokens.

We identify the start token index $j_0$ and employ a timestep-dependent masking to its attention logits:
\begin{equation}
\hat{A}_{i,j_0} =
\begin{cases}
0, & \text{if } t > t_{\text{thresh}} \\[1mm]
A_{i,j_0}, & \text{otherwise}
\end{cases}, \quad i = 1, \dots, L,
\end{equation}
where $t_{\text{thresh}}$ denotes the masking threshold of the timestep in the flow-matching process. The attention to \texttt{<start token>} is masked (set to zero) whenever the timestep exceeds the threshold, otherwise it remains unchanged. 

By reducing the excessive focus on \texttt{<start token>}, the attention distribution becomes more evenly spread across informative text tokens. Consequently, motion frames are encouraged to incorporate richer semantic signals from the entire text sequence, leading to improved fine-grained motion-language alignment.

\begin{figure}[h]
\centering
\includegraphics[width=0.35\textwidth]{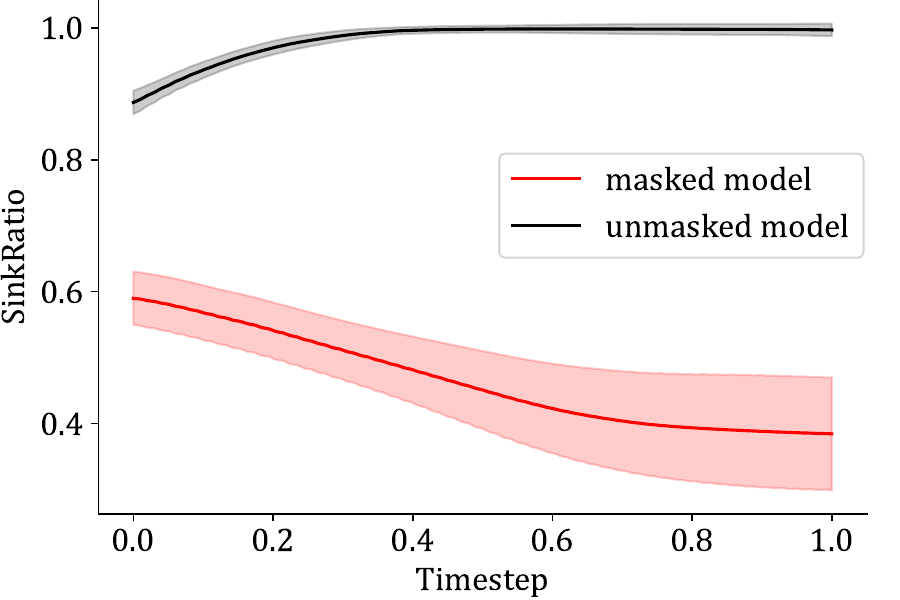}
\caption{SinkRatio curves for masked and unmasked models. Each curve depicts the mean SinkRatio across all batch samples over timesteps, with the shaded region indicating the standard deviation.
}
\label{fig:sinkratio_curves}
\end{figure}

To illustrate the effect of \textit{sink-mask}, we visualize the attention heatmaps of models with \textit{sink-mask} (the masked model) and without \textit{sink-mask} (the unmasked model) under the same textual description and timestep. As shown in Fig. \ref{fig:heatmap_compare}, the unmasked model concentrates nearly all attention weights on \texttt{<start token>}. In contrast, although the masked model still exhibits the attention sink phenomenon, certain informative text tokens maintain relatively high attention weights, such as \texttt{<re>} and \texttt{<arranging>}.

Furthermore, in Fig. \ref{fig:sinkratio_curves}, we plot the SinkRatio curves of both models over increasing timesteps $t$ within the same batch. The unmasked model shows a rising trend, consistently maintaining a high SinkRatio ($0.9\rightarrow 1.0$). In comparison, the masked model exhibits a decreasing trend ($0.6\rightarrow 0.4$), indicating that \textit{sink-mask} effectively mitigates the intensification of attention sink. 

The \textit{sink-mask} mechanism accentuates the informative tokens within textual descriptions, thereby encouraging the model to attend to details with more semantics. Although this process inevitably diminishes the global semantics encoded in \texttt{<start token>}, the global motion priors learnt by the memory slots can mitigate this loss, from where the model can still extract relevant motion knowledge, further underscoring the indispensable role of motion priors in motion-language alignment.

\noindent{\textbf{\textit{Sink-ctrl}: Sink-aware classifier-free guidance. }} Let $X_{\text{cond}}$ and $X_{\text{uncond}}$ denote the conditional and unconditional predictions of flow model. In standard CFG, the conditional and unconditional predictions are combined as
\begin{equation}
X = X_{\text{uncond}} + w(X_{\text{cond}} - X_{\text{uncond}}),
\end{equation}
where $w$ denotes the guidance scale, and the difference $E = X_{\text{cond}} - X_{\text{uncond}}$ represents the original guidance signal. Moreover, in CFG generation, the unconditional branch typically captures only coarse global structures, which can induce generation instability, an effect that is more pronounced in smaller models. To address this, we additionally incorporate local textual features $C_l$ as auxiliary conditioning into $X_{\text{uncond}}$ generation ($C_l$ for the small-scale model and $0.5C_l$ for the big-scale model).

Inspired by CFG-ctrl \citep{wang2026cfg}, we propose \textit{sink-ctrl}, a sink-aware CFG strategy to dynamically regulate guidance according to the attention sink effect. Specifically, the control signal is defined as $S = E + (\lambda_{\text{ctrl}} - 1)\, \hat{E}_\text{prev}$, where $\hat{E}_\text{prev}$ is the regulated guidance from the previous timestep, and $\lambda_{\text{ctrl}}$ controls the strength of temporal coupling. Based on SinkRatio, we compute an adaptive control coefficient
\begin{equation}
k_{\text{eff}} = k_{\text{base}} (1 + \alpha \cdot \text{SinkRatio}),
\end{equation}
where $k_{\text{base}}$ and $\alpha$ are hyperparameters. The guidance is then rectified as $\hat{E} = E - k_{\text{eff}}\cdot \text{sign}(S)$. Finally, the prediction used for sampling is $X = X_{\text{uncond}} + w \cdot \hat{E}$.

\begin{table*}[!t]
\centering
\small
\resizebox{0.9\textwidth}{!}{
\begin{tabular}{lccccccc}
\toprule
\multirow{2}{*}{\textbf{Methods}} 
& \multirow{2}{*}{\textbf{FID}$ \downarrow$} 
& \multicolumn{3}{c}{\textbf{R-Precision} $\uparrow$} 
& \multirow{2}{*}{\textbf{Matching} $\downarrow$} 
& \multirow{2}{*}{\textbf{MModality} $\uparrow$} 
& \multirow{2}{*}{\textbf{CLIP-score} $\uparrow$} \\

\cmidrule(lr){3-5}

&  & \textbf{Top 1} &\textbf{ Top 2} & \textbf{Top 3} &  &  &  \\

\midrule

Real & 0.000$^{\pm.000}$ & 0.503$^{\pm.002}$ & 0.696$^{\pm.001}$ & 0.795$^{\pm.002}$ & 3.244$^{\pm.005}$ & - & 0.639$^{\pm.001}$ \\ \midrule


MotionDiffuse& 0.778$^{\pm.005}$ & 0.450$^{\pm.006}$ & 0.641$^{\pm.005}$ & 0.753$^{\pm.005}$ & 3.490$^{\pm.023}$ & \underline{3.179}$^{\pm.046}$ & 0.606$^{\pm.004}$ \\

ReMoDiffuse & 0.883$^{\pm.021}$ & 0.468$^{\pm.003}$ & 0.653$^{\pm.003}$ & 0.754$^{\pm.005}$ & 3.414$^{\pm.020}$ & 2.703$^{\pm.154}$ & 0.621$^{\pm.003}$ \\

MLD++ & 2.027$^{\pm.021}$ & 0.500$^{\pm.003}$ & 0.691$^{\pm.002}$ & 0.789$^{\pm.001}$ & 3.220$^{\pm.008}$ & 1.924$^{\pm.065}$ & 0.639$^{\pm.002}$ \\

MotionLCM V2 & 2.267$^{\pm.023}$ & 0.501$^{\pm.002}$ & 0.693$^{\pm.002}$ & 0.790$^{\pm.002}$ & 3.192$^{\pm.009}$ & 1.780$^{\pm.062}$ & 0.640$^{\pm.003}$ \\

MARDM-\textbf{$\mathbf{\epsilon}$} & 0.116$^{\pm.004}$ & 0.492$^{\pm.006}$ & 0.690$^{\pm.005}$ & 0.790$^{\pm.005}$ & 3.349$^{\pm.010}$ & 2.470$^{\pm.053}$ & 0.637$^{\pm.005}$ \\

MARDM-\textbf{v} & 0.114$^{\pm.007}$ & 0.500$^{\pm.004}$ & 0.695$^{\pm.003}$ & 0.795$^{\pm.003}$ & 3.270$^{\pm.009}$ & 2.231$^{\pm.071}$ & 0.642$^{\pm.002}$ \\

\midrule

ACMDM-S & 0.107$^{\pm.006}$ & 0.504$^{\pm.003}$ &  0.699*$^{\pm.002}$ &  0.795*$^{\pm.002}$ & 3.258$^{\pm.010}$ & 2.181$^{\pm.085}$ &  0.639*$^{\pm.001}$ \\

\textbf{MLA-Gen-S} & \underline{0.056*}$^{\pm.003}$ & 0.507*$^{\pm.003}$ & 0.698$^{\pm.003}$ & 0.795*$^{\pm.002}$ & 3.237*$^{\pm.011}$ & 2.198*$^{\pm.072}$ & 0.639*$^{\pm.001}$ \\

\midrule

ACMDM-B & 0.083$^{\pm.004}$ & \underline{0.522}$^{\pm.002}$ & \underline{0.717}$^{\pm.003}$ & \underline{0.810}$^{\pm.003}$ & \underline{3.178}$^{\pm.011}$ &  1.790$^{\pm.075}$ & \underline{0.652}$^{\pm.001}$ \\

\textbf{MLA-Gen-B} & \textbf{0.040*}$^{\pm.002}$ & \textbf{0.527*}$^{\pm.002}$ & \textbf{0.721*}$^{\pm.003}$ & \textbf{0.814*}$^{\pm.002}$ & \textbf{3.108*}$^{\pm.008}$ & 1.818*$^{\pm.061}$ & \textbf{0.656*}$^{\pm.001}$ \\

\bottomrule
\end{tabular}
}
\caption{Quantitative text-to-motion evaluation on HumanML3D dataset. We repeat the evaluation 20 times and report the average with 95\% confidence interval. We use bold face/underline to indicate the best/2nd results, and asterisk (*) to indicate the better results between our method and ACMDM \citep{meng2025absolute}.}
\label{tab:results}
\end{table*}

When SinkRatio is large, indicating severe attention concentration, the coefficient $k_{\text{eff}}$ increases, amplifying the control gain applied to the guidance signal. This mechanism enforces stronger corrective updates, which aids the model in achieving stronger semantic alignment. Conversely, when SinkRatio is small, the guidance remains largely unchanged, allowing CFG to exploit informative textual conditions. This strategy helps balance global semantic control with motion-language alignment during generation.

\section{Experiments}
\subsection{Experimental Setup}

\noindent{\textbf{Datasets and Baselines. }}To ensure a fair comparison between MLA-Gen and existing motion generation approaches, we follow the configuration in prior works \citep{meng2025absolute}, adopting the widely-used HumanML3D \cite{guo2022generating} and KIT-ML \citep{plappert2016kit} benchmarks for both model training and evaluation. HumanML3D comprises 14,616 motion sequences, each annotated with multiple textual descriptions, yielding a total of 44,970 text annotations. The dataset is split into training, validation, and test sets with a ratio of 80:15:5. To further verify the generalization capability of our method, we additionally conduct experiments on KIT-ML , following the standard protocol in prior work. 
We compare our method with state-of-the-art motion generation methods, including 
MotionDiffuse~\citep{zhang2024motiondiffuse}, ReMoDiffuse~\cite{zhang2023remodiffuse}, MLD++~\cite{dai2024motionlcm}, MotionLCM V2~\cite{dai2024motionlcm}, MARDM~\citep{meng2025rethinking}, and ACMDM \citep{meng2025absolute}. 

\noindent{\textbf{Implementation details.}} Our model adopts ACMDM~\citep{meng2025absolute} as the backbone architecture. We train variants at small and big scales, denoted as MLA-Gen-S and MLA-Gen-B, respectively corresponding to ACMDM-S and ACMDM-B. 
Hyperparameters are detailed in the supplementary material. All experiments are conducted on a single NVIDIA GeForce RTX 4090 GPU with 24GB memory. The training of MLA-Gen-S takes approximately 8 hours, while MLA-Gen-B requires around 4 days.

\noindent{\textbf{Evaluation protocols.}} For a more comprehensive evaluation, we adopt the evaluator proposed in \citep{meng2025rethinking} for testing and metric computation. We employ Fréchet Inception Distance (\textbf{FID}) to measure the discrepancy between the distributions of generated and real motions, \textbf{R-Precision} (Top-1/2/3), \textbf{Matching Score}, and \textbf{CLIP Score} to assess text-motion alignment, and \textbf{MModality} to quantify the diversity of motions generated from the same textual description. Among these metrics, lower FID and Matching scores, together with higher R-Precision, MModality, and CLIP scores, indicate superior motion generation quality.

\begin{table}[h]
\centering
\small
{
\begin{tabular}{l|cccc}
\toprule
\textbf{Method} & \textbf{FID} $\downarrow$ & \textbf{R-Top1} $\uparrow$ & \textbf{R-Top2} $\uparrow$ & \textbf{R-Top3} $\uparrow$ \\
\midrule
MotionDiffuse & 3.845 & 0.344 & 0.536 & 0.658 \\
ReMoDiffuse & 1.725 & 0.356 & 0.572 & 0.706 \\
MARDM-$\epsilon$ & 0.340 & 0.375 & 0.597 & 0.739 \\
MARDM-v & 0.242 & 0.387 & 0.610 & 0.749 \\
ACMDM & 0.237 & 0.391 & \underline{0.615} & \underline{0.752} \\ 
\midrule
MLA-Gen-S  & \underline{0.199} & \underline{0.392} & \underline{0.615} & 0.744 \\
MLA-Gen-B  & \textbf{0.197} & \textbf{0.395} & \textbf{0.619} & \textbf{0.753} \\
\bottomrule
\end{tabular}
}
\caption{Evaluation results on KIT-ML dataset.}
\label{kit}
\end{table}

\subsection{Experimental Results}

\noindent{\textbf{Quantitative analysis.}} Tn Tab. \ref{tab:results}, We present a quantitative comparison between our method and state-of-the-art text-to-motion generation baselines on HumanML3D. Our method achieves superior performance across several key metrics, including FID ($0.083 \rightarrow 0.040$), R-Precision ($0.522 \rightarrow 0.527$ for Top-1 score), Matching ($3.178 \rightarrow 3.108$), and CLIP Score ($0.652 \rightarrow 0.656$). Compared to existing approaches, MLA-Gen demonstrates a stronger capability to generate high-fidelity motions with accurate semantic alignment. 

Notably, in MLA-Gen-S, the FID score is reduced from 0.107 to 0.056, even surpassing the FID reported for ACMDM-XL (0.058). Meanwhile, in MLA-Gen-B, the performance improvements are more pronounced compared to the S-scale model, particularly in terms of R-Precision (in terms of Top-3 score, $0.810\rightarrow 0.814$ for MLA-Gen-B, while $0.795\rightarrow 0.795$ for MLA-Gen-S). This can be attributed to the increased model capacity, which provides a higher-dimensional attention space for motion-language alignment, facilitating more effective semantic representation. As a result, the generated motions exhibit stronger consistency with the underlying textual semantics.

We further evaluate MLA-Gen on KIT-ML to assess its generalization ability under a different motion distribution. As shown in Tab.~\ref{kit}, MLA-Gen consistently improves upon the baseline across major metrics, confirming the effectiveness of the proposed modules beyond a single dataset.

\begin{figure*}[t]
\centering
\includegraphics[width=\textwidth]{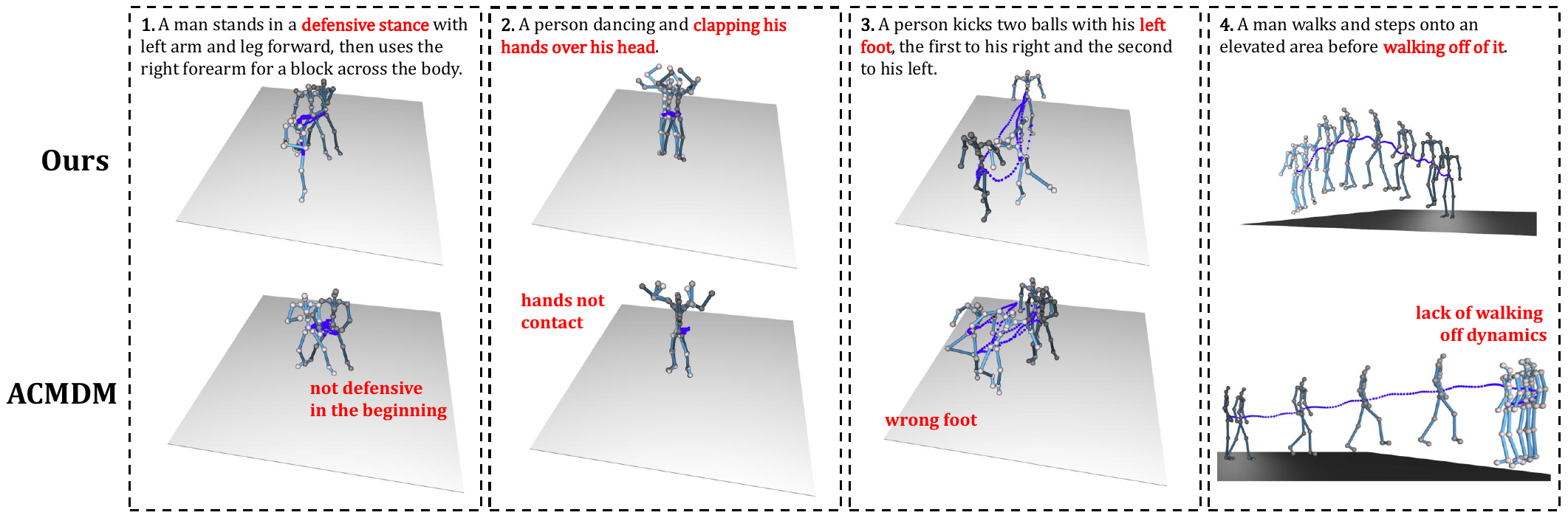}
\caption{Visualization comparison between ACMDM-S \citep{meng2025absolute} and our MLA-Gen-S. }
\label{fig:visualization}
\end{figure*}

\noindent{\textbf{Visualization comparison.}} Fig. \ref{fig:visualization} presents the visualization results of our model MLA-Gen, in comparison with ACMDM \citep{meng2025absolute}. Our approach achieves more precise alignment in the textual descriptions, including joint-level correspondence (e.g., whether the hands are close during clapping in the 2nd group or which foot is used for kicking in the 3th group) as well as temporal consistency (e.g., beginning with a defensive stance in the 1st group and ending with walking off dynamics in the 4th group).

\noindent{\textbf{Human study.}} We conduct a human study to evaluate the perceptual quality and text-motion alignment of MLA-Gen against ACMDM. Specifically, 20 motion pairs are randomly selected and presented in randomized order to 15 participants for evaluation. MLA-Gen is preferred over ACMDM in 78.33\% of cases, indicating that our generated motions are consistently perceived as more realistic and better aligned with the textual descriptions.

\subsection{Ablation Study}

We train and evaluate models without modules we propose to examine their contributions, include the memory slots $M$, the motion-language alignment module $A$, and the sink-based strategy $S$ (\textit{sink-mask} and \textit{sink-ctrl}). For a fair comparison, we employ the original CFG with the same guidance scale in ablations without $S$. 

\begin{table}[H]
\centering
\small

\begin{tabular}{lcccc}
\toprule
\textbf{Methods} & \textbf{FID}$\downarrow$ & \textbf{R-Top1}$\uparrow$ & \textbf{R-Top2}$\uparrow$ & \textbf{R-Top3}$\uparrow$ \\
\midrule
+ $M$ + $A$ + $S$ & \textbf{0.056} & \textbf{0.507} & \textbf{0.698} & \textbf{0.795} \\ \midrule
+ $M$ + $A$ & 0.076 & 0.508 & 0.703 & 0.799 \\ 
+ $M$ & 0.110 & 0.506 & 0.699 & 0.797 \\
+ $A$ & 0.101 & 0.508 & 0.702 & 0.799 \\
w/o any modules & 0.120 & 0.507 & 0.702 & 0.798 \\
\bottomrule
\end{tabular}
\caption{Ablation study of MLA-Gen. We use bold face to denote the original configurations. 
}
\label{tab:ablation_mem}
\end{table}

Tab.\ref{tab:ablation_mem} shows that MLA-Gen achieves strong performance only when global priors from memory slots are combined with fine-grained alignment features, indicating that both components are essential for effective motion-language modeling. Moreover, sink-based strategy suppresses overly dominant global semantics, slightly reducing R-Precision accuracy, but significantly improves FID score. This trade-off stems from the fact that R-Precision primarily evaluates coarse semantic retrieval consistency, whereas our approach focuses on fine-grained token-level grounding. The efficacy of this design choice is further corroborated by SinkRatio, attention distributions, and substantial FID gains.

We provide more ablation studies and sensitivity analyses‌ in the supplementary material, further analyzing the contribution of each component and hyperparameter sensitivity in MLA-Gen.

\section{Conclusion}

In this paper, we explore the role of motion-language alignment in text-driven human motion generation. We introduce MLA-Gen, a framework that explicitly models alignment through global motion priors and fine-grained local conditioning. Furthermore, we identify the attention sink phenomenon in cross-modal alignment and propose SinkRatio as a metric to quantify attention concentration.

Leveraging SinkRatio, we develop alignment-aware generation strategies, including sink-aware token masking and adaptive CFG guidance regulation. These mechanisms dynamically modulate conditional signals based on alignment statistics, enhancing the utilization of textual cues while maintaining global motion coherence. Extensive experiments demonstrate that our approach significantly improves motion quality and semantic consistency.

\noindent \textbf{Limitations and future work.} Despite these advances, several limitations remain. First, the current alignment mechanisms in MLA-Gen rely on attention-based interactions, which may struggle with very long textual descriptions or highly complex motion semantics. Second, while SinkRatio quantifies attention concentration, it does not directly capture higher-order semantic dependencies among tokens. Future research may explore more expressive alignment diagnostics and incorporate structured priors into motion generation models. Another promising direction is extending alignment-aware generation to other multimodal synthesis tasks, such as video generation and embodied agent control.

In conclusion, our findings underscore the importance of understanding and regulating alignment dynamics in multimodal generative models. Our results show that explicitly modeling fine-grained motion-language alignment can improve both motion quality and semantic consistency, while SinkRatio-based strategies further enhance the utilization of textual cues during generation. We believe this perspective will inspire further research on more semantically coherent cross-modal generation.

{
    \small
    \bibliographystyle{ieeenat_fullname}
    \bibliography{main}
}

\clearpage
\setcounter{page}{1}
\maketitlesupplementary

\begin{appendix}

\section{Overview}

In this supplementary material, we provide additional details and results to support the main submission. 

Specifically, 
\begin{itemize}
    \item Section \ref{secA} presents hyperparameters of MLA-Gen.
    \item Section \ref{secB} presents extended ablation studies.
    \item Section \ref{SecC} presents more qualitative visualizations.
\end{itemize}

\section{Hyperparameters} 
\label{secA}

\begin{table}[h]
\centering
\caption{\textbf{Hyperparameter settings.} Hyperparameters listed above the dividing line are those required for training, while those below correspond specifically to the MLA-Gen model.}
\small
\label{tab_hyperparameter}
\begin{tabular}{l l l}
\toprule
\textbf{Name} & \textbf{Meaning} & \textbf{Value} \\
\midrule
ep & Number of training epochs & 500 \\
opt & Training optimizer & AdamW \\
lr & Learning rate & 2e-4 \\
lr-decay & Weight decay & 0.1 \\ \midrule
$S$ & Number of slots & 16 \\
$\lambda$ & Local cond scale & 0.2 \\
top-$K$ & Highest $K$ token in SinkRatio & top-2 \\
$t_\text{thresh}$ & $t$ threshold in \textit{sink-mask} & 0.2 \\
$w$ & Guidance scale in CFG & 4 \\
$\lambda_{\text{ctrl}}$ & Temporal factor in \textit{sink-ctrl} & 6\\
$k_\text{base}$ & Base factor in \textit{sink-ctrl} & 2 \\
$\alpha$ & Amplification factor in \textit{sink-ctrl} & 0.18 \\
\bottomrule
\end{tabular}
\end{table}

\section{Supplementary Ablation Studies}
\label{secB}

We conduct ablation studies from the following five perspectives: (1) memory slots and motion-language alignment; (2) the top-$K$ algorithm in SinkRatio; (3) the \textit{sink-mask} mechanism;  (4) the local conditions in the CFG unconditional branch; and (5) the \textit{sink-ctrl} CFG strategy.

For brevity, all evaluations are performed on MLA-Gen-S, and only the critical metrics, FID and R-Precision, are presented. Unless otherwise specified, all ablated variants share the same settings as the original model, except for the component under investigation.

\subsection{Memory Slots and Motion-Language Alignment}

In the original MLA-Gen, the final textual conditioning $C$ is constructed by integrating the global CLIP features $C_g$ with the local, fine-grained features $C_l$ produced by the cross-modal alignment module, formulated as $C=C_g+\lambda C_l$, where $\lambda$ is fixed at 0.2. To investigate the impact of local feature fusion on overall generation quality, we further train model variants under different settings of $\lambda$, including fixed values of 0.1, 0.4 and 0.6, as well as a learnable parameter optimized during training, then evaluate them using the original CFG with the same guidance scale. Notably, after convergence, the learnable $\lambda$ attains a final value of 0.0011. 

\begin{table}[H]
\centering
\small
\caption{\textbf{Ablation study of the local-to-global fusion factor $\lambda$.} We use \colorbox{gray!20}{gray shade} and \textbf{bold face} to indicate the original configurations in our MLA-Gen.}
{
\begin{tabular}{l|cccc}
\toprule
\multirow{2}{*}{Methods} 
& \multirow{2}{*}{FID$\downarrow$} 
& \multicolumn{3}{c}{R-Precision$\uparrow$}  \\
\cmidrule(lr){3-5}
&  & Top 1 & Top 2 & Top 3  \\
\midrule
$\lambda = 0.1$ & 0.089 & 0.508 & 0.703 & 0.799 \\
\cellcolor{gray!20}\textbf{$\lambda = 0.2$} & \cellcolor{gray!20}\textbf{0.076} & \cellcolor{gray!20}\textbf{0.508} & \cellcolor{gray!20}\textbf{0.703} & \cellcolor{gray!20}\textbf{0.798} \\
$\lambda = 0.4$ & 0.089 & 0.501 & 0.695 & 0.793 \\
$\lambda = 0.6$ & 0.095 & 0.499 & 0.692 & 0.788 \\
learnable $\lambda$ & 0.118 & 0.510 & 0.703 & 0.799 \\ 
\bottomrule
\end{tabular}
}

\label{tab1}
\end{table}

The results are shown in Tab. \ref{tab1}. For fixed $\lambda$ values, 
we observe that generation quality begins to degrade noticeably when $\lambda$ exceeds 0.2, as excessive local conditioning biases the model toward fine-grained cues and diminishes its ability to capture overall semantics, causing it to focus less on holistic understanding. Compared to $\lambda=0.2$, the learnable $\lambda$ converges to a smaller value, with a markedly worse FID score. This occurs because during training, the optimizer favors global features $C_g$ for stability. However, with such a small $\lambda$, local cues are under-utilized, resulting in deviations in fine-grained motion details and higher FID scores.

In Tab.~\ref{ablation}, we additionally analyze the sensitivity to the memory slot number $S$. As $S$ increases from 4/8 to 12, performance improves noticeably, while the differences between $S=12$ and $S=16$ become negligible, suggesting that the memory capacity has largely saturated. This indicates that the method is not sensitive to the exact slot number once sufficient capacity is provided.

\begin{table}[H]
\centering
\small
\caption{\textbf{Ablation study on MLA-Gen-S of slot number $S$.}}
\begin{tabular}{l|cccc}
\toprule
$S$ & FID$\downarrow$ & R-Top1$\uparrow$ & R-Top2$\uparrow$ & R-Top3$\uparrow$ \\

\midrule
 4 & 0.081 & 0.501 & 0.696 & 0.793 \\ 
 8 & 0.081 & 0.503 & 0.696 & 0.791 \\
 12 & 0.056 & 0.507 & 0.697 & 0.796 \\ 
 \cellcolor{gray!20}\textbf{16} & \cellcolor{gray!20}\textbf{0.056} & \cellcolor{gray!20}\textbf{0.507} & \cellcolor{gray!20}\textbf{0.698} &\cellcolor{gray!20}\textbf{ 0.795} \\ 

\bottomrule
\end{tabular}%
\label{ablation}
\end{table}

\subsection{Top-$K$ in SinkRatio}

In motion-language alignment, we find that the top-3 text tokens consistently dominate the attention distribution. Therefore, we analyze the generative quality of MLA-Gen-S with the top-$K$ parameter set from 1 to 3. 

\begin{table}[H]
\centering
\small
\caption{\textbf{Ablation study of the top-$K$ algorithm in SinkRatio.} }
{
\begin{tabular}{l|cccc}
\toprule
\multirow{2}{*}{Methods} 
& \multirow{2}{*}{FID$\downarrow$} 
& \multicolumn{3}{c}{R-Precision$\uparrow$}  \\
\cmidrule(lr){3-5}
&  & Top 1 & Top 2 & Top 3  \\
\midrule
top-1 & 0.051 & 0.506 & 0.698 & 0.793 \\
\cellcolor{gray!20}\textbf{top-2} & \cellcolor{gray!20}\textbf{0.056} & \cellcolor{gray!20}\textbf{0.507} & \cellcolor{gray!20}\textbf{0.698} & \cellcolor{gray!20}\textbf{0.795} \\
 
top-3 & 0.059 & 0.507 & 0.699 & 0.794 \\ 
\bottomrule
\end{tabular}
}
\label{tab2}
\end{table}

As shown in Tab. \ref{tab2}, increasing $K$ improves R-Precision but concurrently degrades FID. This trend arises because a larger $K$ effectively amplifies the control gain applied to the guidance signal: while this reinforces alignment with key tokens, it also perturbs the model’s sampling trajectory, introducing distributional deviations. Notably, Top-2 provides an optimal trade-off, sufficiently leveraging the most informative tokens to enhance alignment while avoiding excessive distortion that would compromise fidelity.

\subsection{\textit{Sink-mask} Mechanism}

\begin{table}[htbp]
\centering
\small
\caption{\textbf{Ablation of the \textit{sink-mask} mechanism.} Strong and weak masks correspond to $t_\text{thresh}=0.2$ and $0.6$.}
\begin{tabular}{lcccc}
\toprule
Methods & FID$\downarrow$ & Top-1$\uparrow$ & Top-2$\uparrow$ & Top-3$\uparrow$ \\
\midrule
\cellcolor{gray!20}\textbf{strong mask} & \cellcolor{gray!20}\textbf{0.056} & \cellcolor{gray!20}\textbf{0.507} & \cellcolor{gray!20}\textbf{0.698} & \cellcolor{gray!20}\textbf{0.795} \\
weak mask & 0.069 & 0.505 & 0.697 & 0.795 \\
no mask & 0.099 & 0.508 & 0.700 & 0.797 \\
\bottomrule
\end{tabular}
\label{tab:ablation_mask}
\end{table}

 In addition to the pre-configured MLA-Gen-S (strong mask, $t_{\text{thresh}}=0.2$), we train and evaluate models with weak mask ($t_{\text{thresh}}=0.6$) and without applying any mask. The results are shown in Tab. \ref{tab:ablation_mask}. We observe that as the intensity of \textit{sink-mask} increases, the FID score exhibits a consistent and notable decrease, whereas the R-Precision remains relatively stable. Accordingly, we select the configuration of more mask with $t_{\text{thresh}}=0.2$ to train our primary model.

\subsection{Local Conditions in the CFG unconditioned branch}

\begin{table}[htbp]
\centering
\small
\caption{\textbf{Ablation of local conditioning in the CFG unconditional branch.} $C_l$, $0.5C_l$, and $0$ indicate full, limited, and no conditional information in $X_\text{uncond}$ generation, respectively.}
\begin{tabular}{lcccc}
\toprule
Methods & FID$\downarrow$ & Top-1$\uparrow$ & Top-2$\uparrow$ & Top-3$\uparrow$ \\
\midrule
\cellcolor{gray!20}\textbf{$C_l$} & \cellcolor{gray!20}\textbf{0.056} & \cellcolor{gray!20}\textbf{0.507} & \cellcolor{gray!20}\textbf{0.698} & \cellcolor{gray!20}\textbf{0.795} \\
$0.5C_l$ & 0.070 & 0.506 & 0.699 & 0.795 \\
$0$ & 0.101 & 0.505 & 0.698 & 0.793 \\
\bottomrule
\end{tabular}
\label{tab:ablation_cond}
\end{table}

To investigate the role of local textual features $C_l$ in the CFG unconditional branch, we further evaluate the model’s performance in generating $X_\text{uncond}$ under two settings: without any conditioning (0), and with limited conditioning ($0.5C_l$). The results in Tab. \ref{tab:ablation_cond} show that introducing $C_l$ on $X_\text{uncond}$ leads to improvements in both FID and R-Precision scores. This suggests that incorporating moderate local constraints into the unconditional branch reduces the distribution gap between conditional and unconditional predictions, resulting to a more stable and discriminative guidance signal. Consequently, the model achieves better text alignment and improved generation quality.

\begin{table*}[!htbp]

\centering
\caption{\textbf{Mapping between video pair paths and their corresponding textual descriptions.} Each pair contains results from ACMDM and our method (MLA-Gen), respectively.}
\label{tab:video_mapping}
\begin{tabular}{p{0.42\linewidth} p{0.52\linewidth}}
\toprule
\textbf{Video Files} & \textbf{Text Description} \\
\midrule

\path{./videos/ACMDM_sample_1.mp4} \newline
\path{./videos/Ours_sample_1.mp4}
&
\textit{A person stands in a defensive stance with the left arm and leg forward, and then uses the right forearm to perform a blocking motion across the body.}
\\ \midrule

\path{./videos/ACMDM_sample_2.mp4} \newline
\path{./videos/Ours_sample_2.mp4}
&
\textit{A person is dancing and clapping their hands above the head.}
\\ \midrule

\path{./videos/ACMDM_sample_3.mp4} \newline
\path{./videos/Ours_sample_3.mp4}
&
\textit{A person kicks two balls with the left foot, first to the right and then to the left.}
\\ \midrule

\path{./videos/ACMDM_sample_4.mp4} \newline
\path{./videos/Ours_sample_4.mp4}
&
\textit{A person walks and steps onto an elevated surface before stepping down.}
\\

\bottomrule
\end{tabular}
\end{table*}

\subsection{\textit{Sink-ctrl} CFG Strategy}

\begin{table}[H]
\centering
\small
\caption{\textbf{Ablation of CFG strategies under different guidance scales.}}
\begin{tabular}{lcccc}
\toprule
Methods & FID$\downarrow$ & Top-1$\uparrow$ & Top-2$\uparrow$ & Top-3$\uparrow$ \\
\midrule
\textit{sink-ctrl} ($w=3.5$) & 0.044 & 0.501 & 0.693 & 0.790 \\
\cellcolor{gray!20}\textbf{\textit{sink-ctrl}} ($w=4$) & \cellcolor{gray!20}\textbf{0.056} & \cellcolor{gray!20}\textbf{0.507} & \cellcolor{gray!20}\textbf{0.698} & \cellcolor{gray!20}\textbf{0.795} \\
\textit{sink-ctrl} ($w=4.5$) & 0.084 & 0.509 & 0.700 & 0.796 \\
\midrule
cfg-ctrl ($w=3.5$) & 0.048 & 0.502 & 0.694 & 0.791 \\
cfg-ctrl ($w=4$) & 0.068 & 0.506 & 0.699 & 0.794 \\
cfg-ctrl ($w=4.5$) & 0.110 & 0.507 & 0.701 & 0.796 \\
\midrule
no ctrl ($w=3.5$) & 0.076 & 0.508 & 0.703 & 0.799 \\
no ctrl ($w=4$) & 0.089 & 0.510 & 0.705 & 0.801 \\
no ctrl ($w=4.5$) & 0.117 & 0.510 & 0.705 & 0.801 \\
\bottomrule
\end{tabular}
\label{tab:ablation_ctrl}
\end{table}

We conduct a detailed comparison of different CFG strategies, including three scheduling approaches: \textit{sink-ctrl}, cfg-ctrl, and original CFG (no ctrl). For each strategy, we evaluate three guidance scales: 3.5, 4, and 4.5. The results are in Tab. \ref{tab:ablation_ctrl}. As the scale increases, the guidance vector strengthens, causing the model to rely more heavily on textual input, which improves R-Precision. However, excessively strong guidance can lead the generated motions to deviate from the real distribution, resulting in FID degrading. Across strategies, \textit{sink-ctrl} consistently outperforms cfg-ctrl on most metrics. Compared to the fixed strategy, \textit{sink-ctrl} suppresses overly dominant global semantics, slightly reducing R-Precision accuracy, but significantly improves FID score.

\section{Supplementary Visualizations}
\label{SecC}

\subsection{Failure Case of MLA-Gen}

\begin{figure}[h]
\centering
\includegraphics[width=0.35\textwidth]{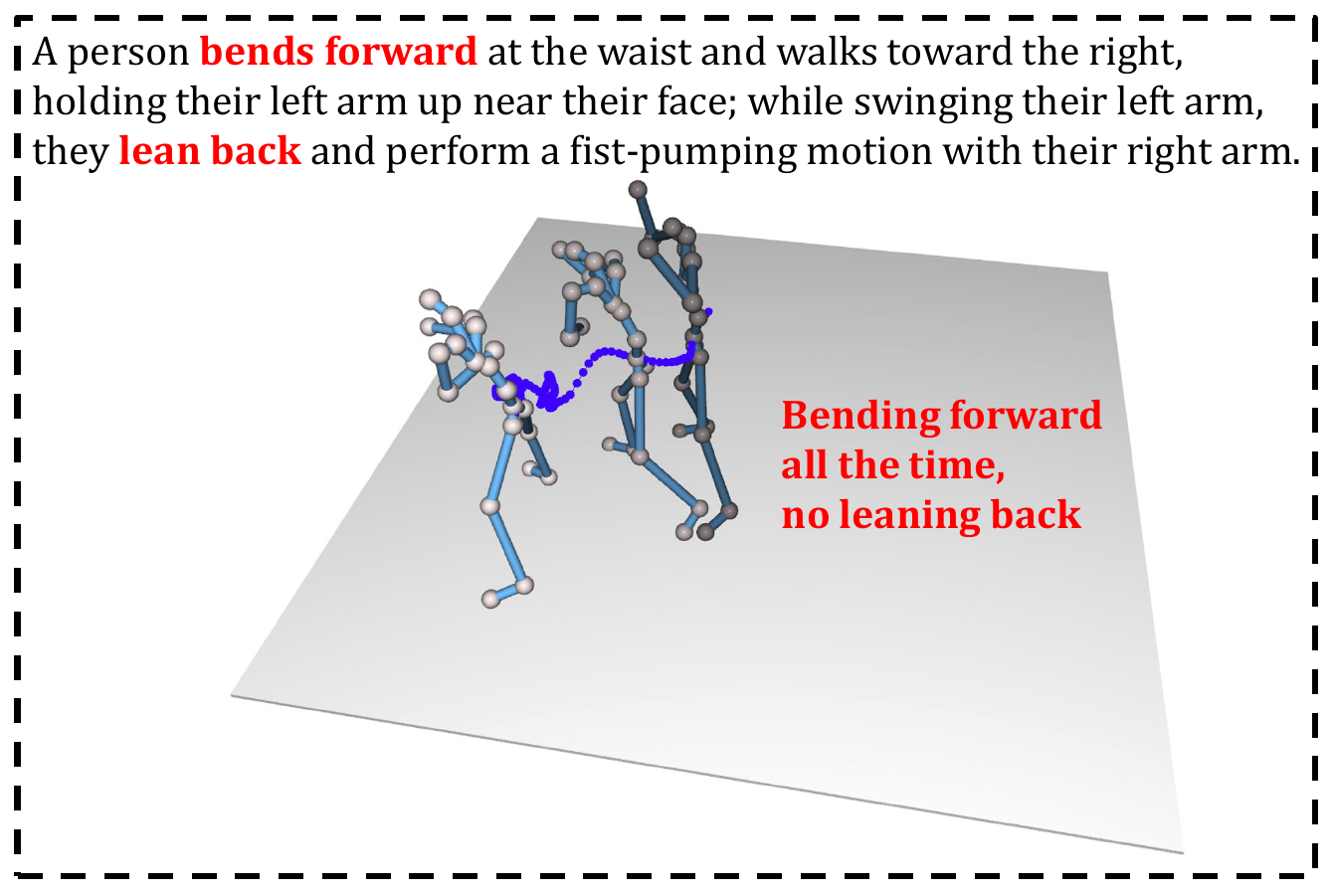}
\caption{\textbf{A failure case of MLA-Gen with a very long textual description.} 
}
\label{mla_fail}
\end{figure}

As illustrated in Fig. \ref{mla_fail}, MLA-Gen relies on attention-based interactions, which may struggle with very long textual descriptions or highly complex motion semantics.

\subsection{Alignment Heatmaps}

\begin{figure*}[t]
\centering
\caption{\textbf{Heatmap sequences of the masked model across increasing timesteps.} Heatmaps are arranged in a left-to-right, top-to-bottom order. Brighter yellow regions indicate higher attention weights between motion frames and text tokens.}
\includegraphics[width=1.0\textwidth]{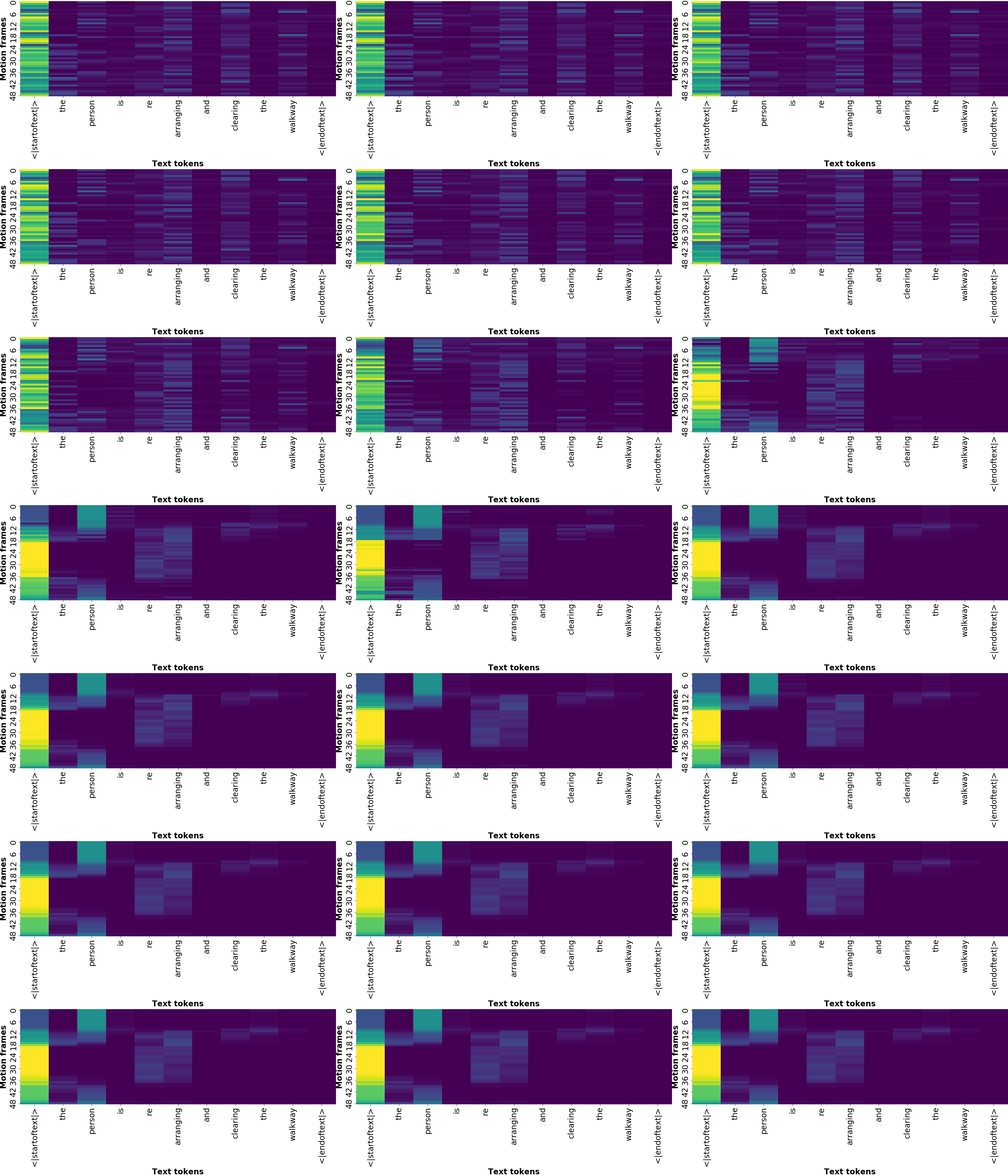}

\label{fig:clip}
\end{figure*}

\begin{figure*}[t]
\centering
\caption{\textbf{Heatmap sequences of the unmasked model across increasing timesteps.}}
\includegraphics[width=1.0\textwidth]{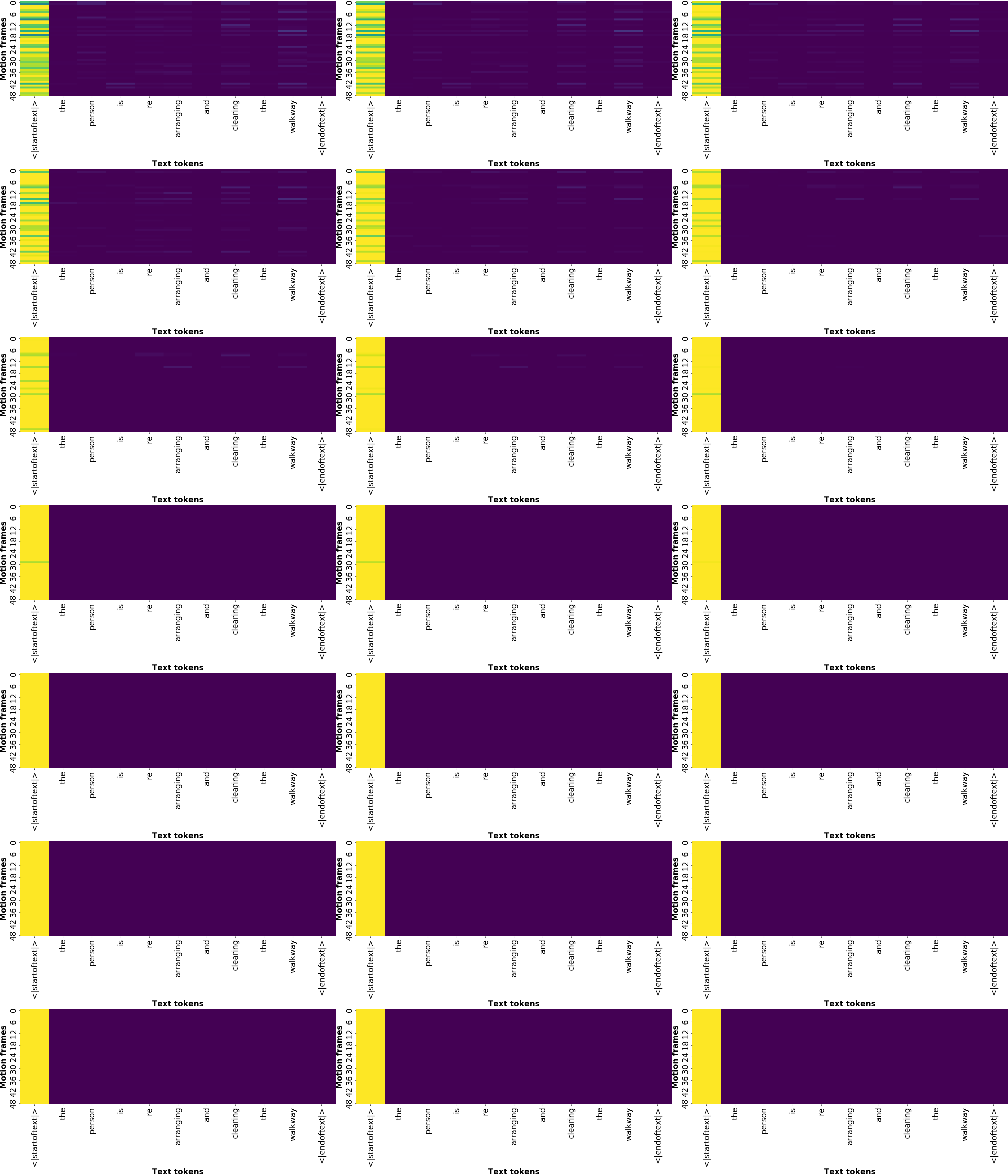}

\label{fig:noclip}
\end{figure*}

To better illustrate motion-language alignment in MLA-Gen and the effect of \textit{sink-mask}, we visualize the evolution of heatmaps over timesteps for both the masked and unmasked models (with and without \textit{sink-mask}). Specifically, we uniformly sample 21 heatmaps (3 rows and 7 columns) across timesteps and arrange them into sequential image grids. Results for the masked model are shown in Fig.~\ref{fig:clip}, and those for the unmasked model in Fig.~\ref{fig:noclip}.

In Fig. \ref{fig:clip}, as the timestep increases, although the attention sink phenomenon emerges at the \texttt{<start token>}, the masked model continues to assign sufficient attention weights to semantically meaningful tokens. In contrast, Fig. \ref{fig:noclip} shows that without \textit{sink-mask}, attention weights become almost entirely concentrated on  \texttt{<start token>}, with the sink effect intensifying over time. These results suggest that MLA-Gen effectively mitigates the attention sink and promotes greater focus on fine-grained textual cues for alignment.

\subsection{Motion Videos}

We provide the full motion videos corresponding to the motion clips used for comparison in the main submission. As PDFs cannot embed playable videos, we include the video filenames for access; reviewers may refer to them in Media Supplement for full inspection. Filenames containing \texttt{"ACMDM"} denote results from the ACMDM model, while those with \texttt{"Ours"} correspond to our model, MLA-Gen. For clarity, each video is accompanied by a caption with its corresponding textual description.

The mapping between filenames and textual descriptions is listed in Tab.~\ref{tab:video_mapping}.

\end{appendix}

\end{document}